# Uncertainty Estimation for Path Loss and Radio Metric Models

Alexis Bose, Jonathan Ethier, Ryan G. Dempsey, and Yifeng Qiu
Communications Research Centre Canada (CRC), Ottawa, Ontario, Canada
{alexis.bose, jonathan.ethier, ryan.dempsey, evanyifeng.qiu}@ised-isde.gc.ca

*Abstract*—This research leverages Conformal Prediction (CP) in the form of Conformal Predictive Systems (CPS) to accurately estimate uncertainty in a suite of machine learning (ML)-based radio metric models [1] as well as in a 2-D map-based ML path loss model [2]. Utilizing diverse difficulty estimators, we construct 95% confidence prediction intervals (PIs) that are statistically robust. Our experiments demonstrate that CPS models, trained on Toronto datasets, generalize effectively to other cities such as Vancouver and Montreal, maintaining high coverage and reliability. Furthermore, the employed difficulty estimators identify challenging samples, leading to measurable reductions in RMSE as dataset difficulty decreases. These findings highlight the effectiveness of scalable and reliable uncertainty estimation through CPS in wireless network modeling, offering important potential insights for network planning, operations, and spectrum management.

## I. INTRODUCTION

Conformal prediction is a versatile framework for uncertainty estimation that applies to any prediction algorithm, offering statistically valid prediction intervals without making assumptions about error distributions. This technique is particularly useful in areas like wireless communications, where assessing prediction uncertainty is essential for effective network planning and management, yet it is currently not widely considered.

Traditional ensemble methods, such as DeepMind's GenCast [3], which involve training multiple models and aggregating their outputs to model the probability distribution of different future scenarios, offer reliable uncertainty estimates. However, these approaches are often computationally intensive, limiting their practicality for large-scale or real-time applications. In contrast, conformal prediction systems (CPS) provide an efficient alternative by requiring only a single model training phase, followed by calibration using a hold-out dataset.

Another technique involves training a model to generate uncertainty outputs, as demonstrated in uncertainty-aware Reference Signal Received Power (RSRP) predictions. [4]. Although this approach demonstrates effective performance, it may produce uncalibrated uncertainty estimates. This limitation arises because the method presumes the model is well-specified—a condition that is frequently unmet in practical scenarios where the necessary knowledge is rarely available, as shown and remedied by conformal prediction in [5].

In this study, we estimate uncertainty by employing CPS within the Conformal Classifiers Regressors and Predictive Systems (CREPES) framework [6], utilizing various difficulty estimators, including Target Strangeness [7]. Specifically, we design a unique Communications Research Centre (CRC)-CPS for each model in [1]: CRC-ML-RSRP, CRC-ML-RSRQ, and CRC-ML-RSSI. The CRC-CPS models each generate difficulty scores and corresponding 95% confidence prediction intervals (PIs). Additionally, we extend this approach to a 2-D map-based ML path loss model (hereafter referred to as map-based CRC-MLPL) [2], using internal features as outlined in [8]. We assess the effectiveness of these intervals in terms of effective coverage and examine the CPS's ability to generalize across different cities. Furthermore, we evaluate the accuracy of the difficulty estimators by measuring the underlying models' root mean squared error (RMSE) on samples of increasing difficulty.

## II. PROPOSED METHOD

The unique CPSs were created using the CREPES framework [6]. The CP training dataset was utilized to calibrate the difficulty estimator, which was then applied to the CP calibration set to obtain refined difficulty measures. Subsequently, a conformal regressor (CR) model was developed by combining these calibrated difficulty estimates with the calibration residuals. This CR model transforms point predictions into PIs at specified confidence levels. For the test dataset, PIs were generated by first deriving difficulty estimates from the predicted targets. These difficulty measures, along with the desired confidence level, were then input into the CR model to produce the final PIs.

In the CREPES framework, normalized conformal regressors implemented within the CPSs include *norm_std*, which standardizes the target variables based on their K-nearest neighbours' standard deviations, and additionally *norm_targ_strng*, the Target Strangeness estimator that was integrated into this framework as introduced in [7]. Additionally, to further explore PIs of varying sizes, the object space was segmented into distinct, groupings of predictions into specific bins, known as Mondrian categories. The Mondrian-based regressors standardize based on the residuals from the K-nearest neighbours, and include *norm_std*, *norm_targ_strng*, and *norm_res*.

The CPS hyperparameter optimization process was applied across the five distinct difficulty estimators and five different random seeds to ensure the robustness of our findings. For each estimator, the combination of hyperparameters that resulted in the narrowest mean PI widths was selected. We evaluated the effective coverage, defined as the proportion of targets that fall within the PIs, to assess the reliability of the PIs. To maintain consistency and comparability, our analyses focused on achieving approximately 90% effective coverage. Consequently, PIs were generated for both 90% and 95% confidence levels. The optimal CPS was identified based on achieving the lowest mean PI width while maintaining the highest effective coverage.

Once the best CPS was identified it was used to create PIs as shown below:

```
#estimate the difficulty, sigmas:
df_test['sigmas'] =
cps_obj['difficulty_estimator'].apply(X=
    df_test[feat_cols],y=df_test['pred'])

#prediction intervals, using sigmas
#and bins if a mondrian CPS
intervals = cps_obj['cps'].predict(df_test
    ['pred'], sigmas=df_test['sigmas'],
    bins=bins_test,lower_percentiles=2.5,
    higher_percentiles=97.5)
```

The full implementation of this method can be found in [9].

## III. DATA PREPARATION

Our study utilizes the dataset introduced in [1], which presents a suite of CRC machine learning models targeting 4G radio metrics: RSRP, RSRQ, and RSSI. These models achieved RMSE performances below 11.69 dB for RSRP, 3.23 dB for RSRQ, and 10.36 dB for RSSI, evaluated on a dataset comprising over 300,000 data points collected from the Toronto, Montreal, and Vancouver areas.

For the purposes of this paper, a Conformal Prediction (CP) calibration set was held out prior to the training of the CRC-ML-4G Radio Metric models. We employed the inductive, or split, CP approach using external tabular features to estimate uncertainty. Specifically, calibration sets for the CRC-CPS-4G Radio Metric models were derived from data pertaining to Toronto, focusing on the geohash6 area dpz833, where a geohash6 represents a 6-character alphanumeric string that defines a geographic area with an approximate precision of 0.61 km by 0.61 km. Details of the datasets used for CRC-CPS-4G Radio Metrics are summarized in Table I.

TABLE I
CRC-CPS-4G RADIO METRICS (RSRP, RSRQ, RSSI) CROWDSOURCED
DATASETS FOR UNCERTAINTY ESTIMATION

| Dataset | Counts Total (Outdoor + Indoor) | Features* | Description |
|---|---|---|---|
| Train | 210,944 (128,464 + 82,480) | 12 or 14 | Toronto** |
| Calibration | 1,293 (1,050 + 243) | 12 or 14 | dpz833 |
| CPS Test | 198,075 (122,463 + 75,612) | 12 or 14 | Vancouver |
| Blind Test | 165,613 (110,001 + 55,612) | 12 or 14 | Montreal |

*Twelve features for RSRP, and fourteen for RSRQ and RSSI.
**Without dpz833, a geohash6 area in downtown Toronto.

The map-based CRC-MLPL is described in [2] and utilizes convolutional neural networks to automatically extract features from high-resolution 2-D obstruction height maps to directly predict path loss. This model was trained using United Kingdom (UK) drive test data and achieves a mean RMSE of 7.35 dB. For this paper, a CP calibration set was held out prior to the training of the map-based CRC-MLPL model. The inductive, or split, CP approach to multimodal regression as demonstrated in [8] leverages internal features, and in this case from the output of the final Convolution + ReLU layer, resulting in 16,384 internal features. The CP calibration set for the map-based CRC-MLPL was a randomly sampled 20% from select UK cities (Boston, London, Merthyr Tydfil, Nottingham and Southampton). The CPS test set was from the city of Stevenage, as detailed in Table II.

TABLE II
CPS DATASETS FOR MAP-BASED CRC-MLPL UNCERTAINTY
ESTIMATION

| Dataset | Counts Total | Features | Description |
|---|---|---|---|
| Train | 7,200 | 16,384 | random 60% from UK cities* |
| Calibration | 4,800 | 16,384 | random 20% from UK cities* |
| CPS Test | 2,400 | 16,384 | random 50% from Stevenage |

*Boston, London, Merthyr Tydfil, Nottingham, and Southampton.

## IV. RESULTS[1]

The uncertainty estimates demonstrated high adaptability, with PIs adjusting their width based on the assessed difficulty of each prediction. This adaptability is illustrated in Fig. 1, which also depicts a box plot showing the distribution of widths corresponding to different prediction difficulties. More exemplary plots can be seen in the Appendix.

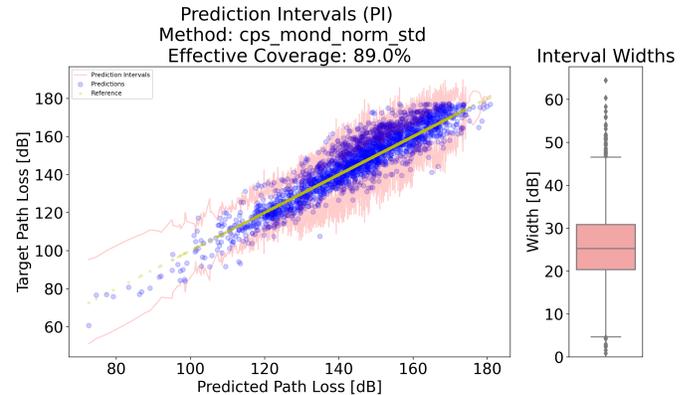

Fig. 1. An example of PIs generated from the CPS map-based CRC-MLPL method is illustrated, along with a box plot showing the distribution of PI widths. Note that it is challenging to visually distinguish the blue prediction densities and the PIs that do not encompass the target values.

One of the primary objectives of any predictive model is its ability to generalize across diverse environments. The CRC

---
[1]The results were processed on an Amazon Web Services m6i.4x-16xlarge EC2 instance.

Radio Metrics' best-performing CPS models were trained on Toronto data and subsequently tuned on data from Vancouver. Additionally, Montreal data was utilized to evaluate the generalization of the tuned CPS models. The results, presented in Table III, indicate that all models demonstrate similar performance levels in both Vancouver and Montreal, thereby confirming their strong generalization capabilities.

Furthermore, the quality of the difficulty estimates was validated by calculating the underlying models' RMSE on samples of increasing difficulty that were progressively excluded from the training set, and can be seen in Table III. This assessment confirmed that the difficulty estimators effectively identified samples with high prediction error and low model confidence, as evidenced by the decrease in RMSE as more difficult samples were held out.

This difficulty insight can now be utilized to identify whether the performance issues arise from the data or the model itself. From the data perspective, difficult samples can be grouped to determine if their distribution is under-represented in the training dataset. These groups may share common feature characteristics, such as being part of indoor or outdoor environments or representing unlikely scenarios. Identifying these difficult data groups helps to quantify the model's limitations, allowing it to avoid making predictions in these difficult regions to maintain desired performance. However, if these difficult data groups are critical to the model's requirements, further focused learning could be employed to enhance the model's ability to handle such cases. This would involve an additional training phase with an emphasis on incorporating more of these difficult data groups. This approach can improve performance while offering partial explainability by identifying data groups that contribute to difficulty.

Finally, the individual PI performance was examined for the map-based CRC-MLPL using the easiest test quartile consisting of 502 samples. The minimum, median, and maximum PIs are in Table IV. The PIs adapt to difficulty and this can be observed by the range of the PI widths. All of the intervals are asymmetric as the CPS is minimizing the PI widths. The effective coverage shows that the PI does not always include the target, and can be seen by the smallest individual PI presented in Table IV.

TABLE III
RMSE on Least Difficult Quartiles (Cumulative) for Various Models and Datasets

| Model Configuration | Test Dataset | RMSE on least difficult (cumulative quartiles) | | | |
|---|---|---|---|---|---|
| | | 25% | 50% | 75% | All |
| CRC-CPS-4G-RSRP | Vancouver | 10.19 | 10.29 | 10.37 | 10.44 |
| | Montreal | 9.37 | 9.42 | 9.56 | 9.76 |
| CRC-CPS-4G-RSRQ | Vancouver | 3.25 | 3.23 | 3.22 | 3.23 |
| | Montreal | 3.10 | 3.12 | 3.14 | 3.18 |
| CRC-CPS-4G-RSSI | Vancouver | 9.68 | 9.94 | 10.18 | 10.36 |
| | Montreal | 8.85 | 9.15 | 9.30 | 9.50 |
| CPS for map-based CRC-MLPL | Stevenage, UK | 6.82 | 6.95 | 6.93 | 7.15 |

TABLE IV
Individual PI results for map-based CRC-MLPL using the easiest test quartile (502 samples)

| Individual PI | Prediction [dB] | PI [L, U] [dB] | PI Width [dB] | Target [dB] |
|---|---|---|---|---|
| Smallest* | 174.8 | [174.3, 175.1] | 0.83 | 175.3 |
| Median* | 158.2 | [150.3, 164.3] | 13.9 | 154.0 |
| Largest* | 169.3 | [157.6, 178.3] | 20.67 | 167.5 |

*Note: intervals are asymmetric.*

## V. Conclusions and Next Steps

Our study demonstrates that the uncertainty estimates obtained using CPS within the CREPES framework are highly adaptable, as evidenced by the variable widths of PIs tailored to the difficulty of each prediction. The models exhibit strong generalization capabilities, maintaining consistent performance across different cities. Moreover, the difficulty estimators accurately identify challenging samples, facilitating reduced prediction errors in future model inference by effectively recognizing and managing prediction uncertainty. These findings suggest that such methodologies can significantly enhance modeling performance by highlighting commonalities within difficult samples and guiding data collection or prediction limits. This advancement also contributes to the pathway towards partial explainability in predictive models.

## VI. Acknowledgment

The authors extend their gratitude to their CRC colleagues for their valuable contributions, expert insights, and support throughout this forward-looking research project.

## VII. APPENDIX

The following plots show how the CPS configurations were chosen and all resulting PIs. Recall the optimal CPS was identified based on achieving the lowest mean PI width while maintaining at least 89% effective coverage.

### A. CRC-CPS-4G-RSRP

The *cps_norm_std* was the only CPS with an effective coverage 89% or higher in Vancouver, and the same configuration was even higher in Montreal.

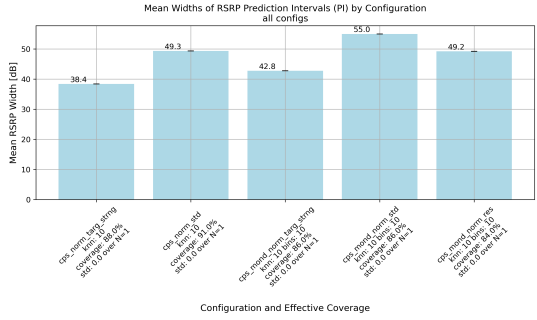

Fig. 2. CRC-CPS-4G-RSRP: Vancouver.

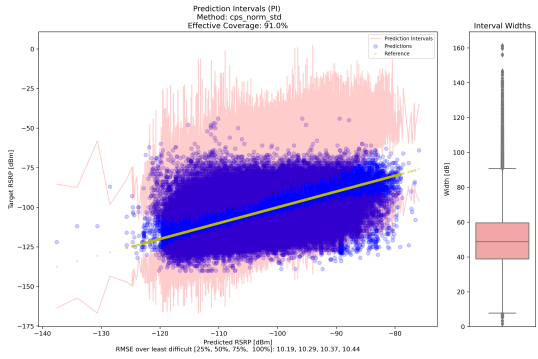

Fig. 3. CRC-CPS-4G-RSRP: Vancouver.

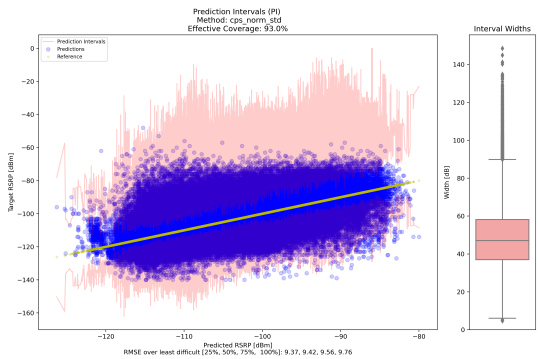

Fig. 4. CRC-CPS-4G-RSRP: Montreal.

### B. CRC-CPS-4G-RSRQ

The *cps_mond_norm_std* had the lowest mean RSRQ PI width and an effective coverage 89% or higher in Vancouver and the same configuration was equivalent in Montreal.

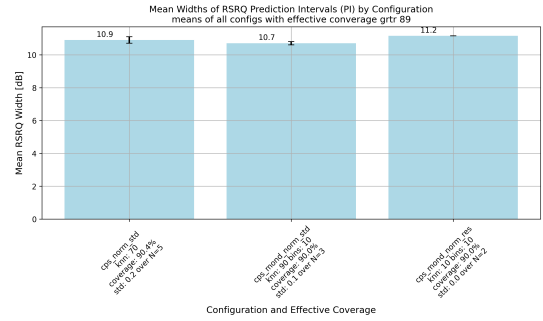

Fig. 5. CRC-CPS-4G-RSRQ: Vancouver.

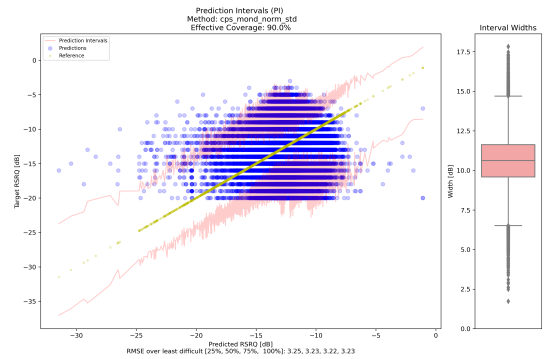

Fig. 6. CRC-CPS-4G-RSRQ: Vancouver.

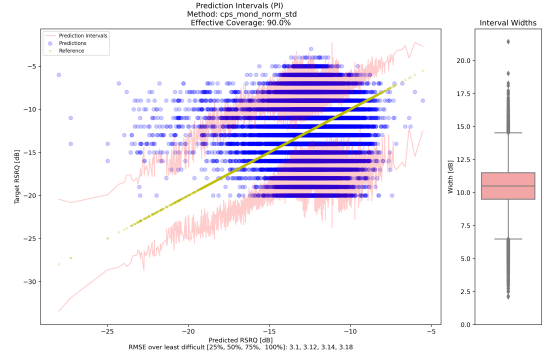

Fig. 7. CRC-CPS-4G-RSRQ: Montreal.

## C. CRC-CPS-4G-RSSI

The *cps_mond_norm_targ_strng* had the lowest RSSI PI width and an effective coverage 89% or higher in Vancouver and the same configuration was even higher in Montreal.

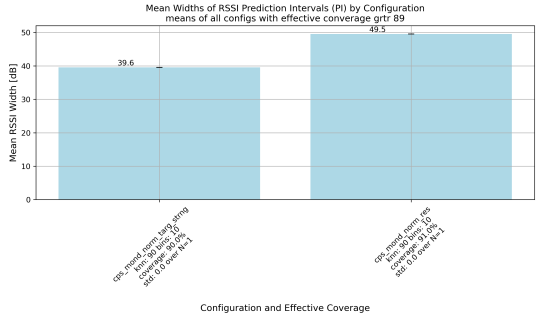

Fig. 8. CRC-CPS-4G-RSSI: Vancouver.

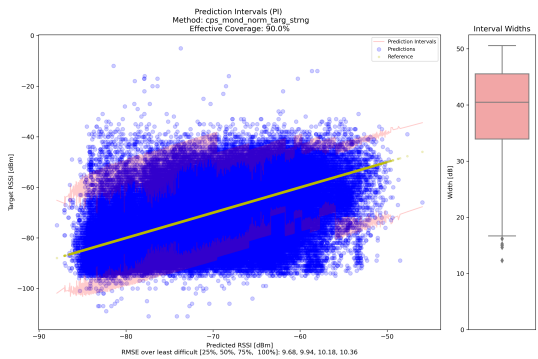

Fig. 9. CRC-CPS-4G-RSSI: Vancouver.

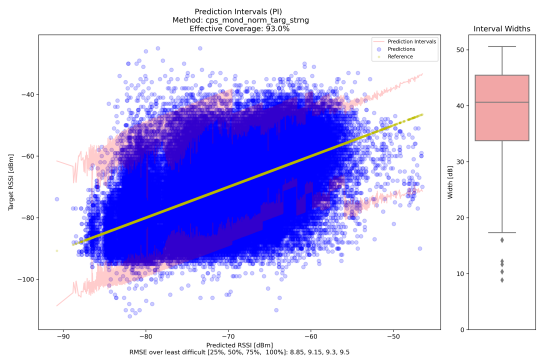

Fig. 10. CRC-CPS-4G-RSSI: Montreal.

## D. CPS for map-based CRC-MLPL

The *cps_mond_norm_std* had the lowest mean path loss width and an effective coverage 89% or higher in Stevenage.

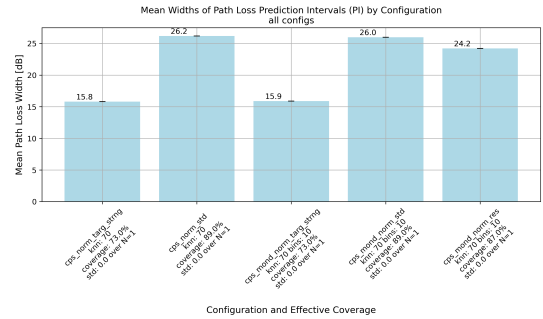

Fig. 11. map-based CRC-MLPL: Stevenage, UK.

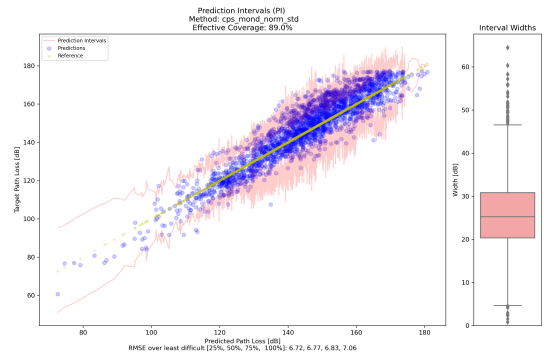

Fig. 12. map-based CRC-MLPL: Stevenage, UK.